# Unsupervised word sense disambiguation in dynamic semantic spaces

**Jean-François Delpech**
WebGlyphs, Inc.
1515 N. Colonial Ct
Arlington, VA 22209-1439
United States
jfdelpech.webglyphs@gmail.com

In this paper, we are mainly concerned with the ability to quickly and automatically distinguish word senses in dynamic semantic spaces in which new terms and new senses appear frequently. Such spaces are built "on the fly" from constantly evolving data sets such as Wikipedia, repositories of patent grants and applications, or large sets of legal documents for Technology Assisted Review and e-discovery. This immediacy rules out supervision as well as the use of *a priori* training sets. We show that the various senses of a term can be automatically made apparent with a simple clustering algorithm, each sense being a vector in the semantic space. While we only consider here semantic spaces built by using random vectors, this algorithm should work with any kind of embedding, provided meaningful similarities between terms can be computed and do fulfill at least the two basic conditions that terms which close meanings have high similarities and terms with unrelated meanings have near-zero similarities.

## 1. Introduction

Word sense disambiguation (WSD) is essential to sentence understanding and is thus a core natural language processing (NLP) problem, since many terms (words or phrases) are polysemous. For example, in Information Retrieval (IR) and Technology Assisted Review (TAR), a query including the word *pig* will have very different interpretations if the word refers to an animal of genus *Sus*, to *pig iron*, to the *Pipeline Inspection Gauge* of the oil industry or to the *picture-in-guide* familiar to television viewers.

As pointed out by Yarowsky [1], the word *disambiguation* may itself have two meanings, either "[…] assigning each instance of a word to established sense *definitions* (such as in a dictionary) […]" or " […] sense *induction*: using distributional similarity to partition word instances into clusters […]".

In this publication, we are mainly concerned by the problem of querying open, constantly evolving data sets such as Wikipedia, repositories of patent grants and applications, or large sets of legal documents for Technology Assisted Review and e-discovery. New words and new senses appear in a seemingly random way after each update, daily or weekly, and any *a priori* human-involved training or classification is unfeasible since significant words may number in millions.

In consequence, the present work will exclusively understand *disambiguation* as equivalent to *sense induction*. This clearly rules out a number of approaches to WSD [2, 3], such as supervised or semi-supervised methods which rely on existing sense-tagged corpora or which use some labeled data to train an initial classifier.

For example, in a recent publication, Butnaru, Ionescu and Hristes [4] propose a new approach inspired by DNA sequencing but "[their] goal is still to find a configuration of senses for the whole document that matches the ground-truth configuration produced by human annotators". For this comparison, they use either WordNet [5] or the pretrained word sense embedding generated by the *word2vec* toolkit [6, 7] from the Google News data set, but static training is obviously not suitable for dynamically evolving data sets.

Yates and Etzioni [8] give an interesting discussion and review of unsupervised disambiguation, where neither hand-tagged training examples nor domain knowledge are available. Their approach may at first glance seem similar to ours as it also relies on clustering and on the *distributional hypothesis*, which can be restated as Firth's Law, "you shall know a word by the company it keeps" [9]. However, their goal is quite different as they aim to extract synonyms (e.g., *Mars* and *Red Planet*) while we seek to identify the various senses of the word (*mars* as a *god*, a *planet* or a *sweet*.)

## 2. Word embeddings and random vectors

Many NLP systems rely on the vector space model introduced by Salton [10, 11], in which each word occupies a separate dimension of a space of very high dimensionality equal to the number of distinct words. In this approach, now often called *one-hot encoding*, the scalar products of vectors associated with distinct words are by definition zero.

To obtain distances (or similarities) between pairs of words, a frequently used NLP technique involves mapping words or phrases from a corpus of documents into vectors of real numbers. In other words, a very high dimensionality space with one dimension per distinct word is embedded into a continuous vector space $\mathbb{R}^d$ of much lower dimensionality $d$. The similarity of two distinct words is then defined as the scalar product of their normalized vectors in $\mathbb{R}^d$ and it ranges from 1.0 (identity) to -1.0 (diametrical opposite.) In most cases, their similarities will be around 0 as words usually have only a small, finite numbers of semantic neighbors.

Many embedding approaches have been considered, from Latent Semantic Indexing [12] to neural networks (see e.g. [6, 7]), with results depending both (i) on the initial corpus (corpora centered on the oil industry, on animal husbandry or on the entertainment industry will yield very different neighbors to the word *pig*), and (ii) on the details of the embedding transformation: for example, is the algorithm linear or not? does it take into account words collocated within a sentence or within a fixed-size window?

A very simple embedding transformation [13, 14, 15, 16] relies on the fact that in $\mathbb{R}^d$ one can create an exponentially large number $N \gg d$ of random vectors quasi-orthogonal to each other [17, 18, 19]. These vectors can then be treated to a good approximation as if they were orthogonal and

a. to each distinct, significant term $t$ in a large set of documents is associated a normalized random *seed vector* belonging to $\mathbb{R}^d$ which is quasi-orthogonal to any other seed vector, and
b. to each $t$ is then attached as *term vector* a linear, weighted combination of the seed vectors of the terms co-occurring with $t$ in all windows of fixed size or in all sentences.

Each term vector $\mathcal{T}_i$ as well as any linear combination of term vectors such as documents are themselves obviously embedded in $\mathbb{R}^d$. In this $d$-dimensional Euclidean semantic space, the similarity $\sigma_{ij}$ between terms $t_i$ and $t_j$ is the scalar product of the associated, normalized term vectors:

$$\sigma_{ij} = \langle \mathcal{T}_i | \mathcal{T}_j \rangle \tag{1}$$

It is sometimes more convenient to consider the distance $\mathcal{D}_{ij}$ which is related to the similarity by $\mathcal{D}_{ij} = \sqrt{2(1-\sigma_{ij})}$.

Distances satisfy the triangle inequality and range from $0$ (same $t_i$ and $t_j$) to $2$ (diametrically opposed vectors,

never found in practice because in high dimension a space is almost everywhere empty, whatever the number of words [17].)

A marked advantage of this linear, transparent process is that it is by nature incremental: a small addition to (or deletion from) the data set involves only a small, finite number of words, at least to a first, very good approximation (over time, word frequencies and thus weight factors will vary), and is thus extremely fast.

In this work, the embedding space dimensionality was $d = 300$ and the window size was $ws = 11$. The random nature of the seed vectors adds some noise to the similarity but this is negligible as the noise's average value is $0$ and its standard deviation is $\sqrt{1/d} \approx 0.058$ for $d = 300$ [18]. About $814,000$ patent applications have been downloaded from the semi-official USPTO site at `http://patentscur.reedtech.com/` between June 2014 and June 2017; these applications cover a broad range of technical categories and the data set contains a total number of over two billion words and well over one million distinct, semantically significant words. The patent data set is quite 'noisy' in that a number of less-frequent words are typos and this does sometimes lead to spurious results.

## 3. Disambiguation algorithm

In a previous article[19], it was shown that the word *mantle* had at least two very different meanings in the patent database: it may refer to a common laboratory equipment, a *heating mantle*, often associated with a *stirrer*, or it may refer to a *mantle cell*, often associated in cancerology with *Burkitt lymphoma*. The following table shows the symmetric similarity matrix for some words associated with *mantle* in the patent database:

|    |                    | 0    | 1    | 2    | 3    | 4    | 5    | 6    | 7    | 8    | 9    | 10   | 11   | 12   | 13   | 14   | 15   | 16   | 17   |
|----|--------------------|------|------|------|------|------|------|------|------|------|------|------|------|------|------|------|------|------|------|
| 0  | mantle             | **1.00** | 0.58 | 0.53 | 0.52 | 0.51 | 0.49 | 0.48 | 0.47 | 0.46 | 0.45 | 0.45 | 0.44 | 0.43 | 0.43 | 0.43 | 0.42 | 0.41 | 0.40 |
| 1  | stirrer            | 0.58 | **1.00** | 0.06 | 0.82 | 0.78 | 0.88 | 0.51 | 0.04 | 0.73 | 0.69 | 0.51 | 0.79 | 0.56 | 0.09 | 0.66 | 0.05 | 0.06 | -0.00 |
| 2  | mcl                | 0.53 | 0.06 | **1.00** | 0.06 | 0.06 | 0.08 | 0.14 | 0.75 | 0.08 | 0.00 | 0.11 | 0.11 | 0.08 | 0.71 | 0.04 | 0.72 | 0.81 | 0.68 |
| 3  | liebig             | 0.52 | 0.82 | 0.06 | **1.00** | 0.55 | 0.74 | 0.48 | 0.01 | 0.51 | 0.64 | 0.59 | 0.62 | 0.53 | 0.10 | 0.48 | 0.07 | 0.06 | 0.03 |
| 4  | gas-adapter        | 0.51 | 0.78 | 0.06 | 0.55 | **1.00** | 0.67 | 0.48 | 0.05 | 0.69 | 0.76 | 0.31 | 0.59 | 0.45 | 0.08 | 0.82 | 0.02 | 0.05 | -0.02 |
| 5  | thermometer        | 0.49 | 0.88 | 0.08 | 0.74 | 0.67 | **1.00** | 0.38 | 0.05 | 0.75 | 0.55 | 0.42 | 0.77 | 0.53 | 0.11 | 0.57 | 0.07 | 0.08 | 0.05 |
| 6  | vigreux            | 0.48 | 0.51 | 0.14 | 0.48 | 0.48 | 0.38 | **1.00** | 0.08 | 0.33 | 0.50 | 0.28 | 0.28 | 0.28 | 0.07 | 0.42 | 0.11 | 0.07 | 0.13 |
| 7  | immunocytoma       | 0.47 | 0.04 | 0.75 | 0.01 | 0.05 | 0.05 | 0.08 | **1.00** | 0.05 | 0.02 | 0.07 | 0.07 | 0.03 | 0.72 | 0.00 | 0.72 | 0.82 | 0.57 |
| 8  | nitrogen-inlet     | 0.46 | 0.73 | 0.08 | 0.51 | 0.69 | 0.75 | 0.33 | 0.05 | **1.00** | 0.40 | 0.36 | 0.65 | 0.47 | 0.10 | 0.56 | 0.06 | 0.06 | 0.03 |
| 9  | round-bottom       | 0.45 | 0.69 | 0.00 | 0.64 | 0.76 | 0.55 | 0.50 | 0.02 | 0.40 | **1.00** | 0.23 | 0.37 | 0.32 | 0.04 | 0.67 | 0.01 | 0.03 | -0.02 |
| 10 | four-paddle        | 0.45 | 0.51 | 0.11 | 0.59 | 0.31 | 0.42 | 0.28 | 0.07 | 0.36 | 0.23 | **1.00** | 0.47 | 0.36 | 0.12 | 0.23 | 0.07 | 0.09 | -0.04 |
| 11 | nitrogen-blowing   | 0.44 | 0.79 | 0.11 | 0.62 | 0.59 | 0.77 | 0.28 | 0.07 | 0.65 | 0.37 | 0.47 | **1.00** | 0.38 | 0.13 | 0.46 | 0.07 | 0.09 | 0.00 |
| 12 | mantel             | 0.43 | 0.56 | 0.08 | 0.53 | 0.45 | 0.53 | 0.28 | 0.03 | 0.47 | 0.32 | 0.36 | 0.38 | **1.00** | 0.15 | 0.40 | 0.13 | 0.11 | 0.05 |
| 13 | angioimmunoblastic | 0.43 | 0.09 | 0.71 | 0.10 | 0.08 | 0.11 | 0.07 | 0.72 | 0.10 | 0.04 | 0.12 | 0.13 | 0.15 | **1.00** | 0.02 | 0.67 | 0.86 | 0.56 |
| 14 | multi-necked       | 0.43 | 0.66 | 0.04 | 0.48 | 0.82 | 0.57 | 0.42 | 0.00 | 0.56 | 0.67 | 0.23 | 0.46 | 0.40 | 0.02 | **1.00** | -0.01 | 0.01 | 0.00 |
| 15 | waldenstrom        | 0.42 | 0.05 | 0.72 | 0.07 | 0.02 | 0.07 | 0.11 | 0.72 | 0.06 | 0.01 | 0.07 | 0.07 | 0.13 | 0.67 | -0.01 | **1.00** | 0.78 | 0.67 |
| 16 | burkitt            | 0.41 | 0.06 | 0.81 | 0.06 | 0.05 | 0.08 | 0.07 | 0.82 | 0.06 | 0.03 | 0.09 | 0.09 | 0.11 | 0.86 | 0.01 | 0.78 | **1.00** | 0.60 |
| 17 | waldenstroms       | 0.40 | -0.00 | 0.68 | 0.03 | -0.02 | 0.05 | 0.13 | 0.57 | 0.03 | -0.02 | -0.04 | 0.00 | 0.05 | 0.56 | 0.00 | 0.67 | 0.60 | **1.00** |

Table 1 - Similarity matrix of words associated with *mantle* in the patent database. While all the words in the cohort have comparable similarities to *mantle*, it can be seen that the similarity between pairs of words can be much lower: for example, *immunocytoma* and *stirrer* or *burkitt* and *liebig* are almost orthogonal to each other. It can be inferred from the table and a list of abbreviations that the highly polysemous term *mcl* stands for *Mantle Cell Lymphoma*, a form of cancer.

A cursory examination of Table 1 suggests a very simple algorithm to disambiguate term $t$ in a completely automatic way, using only the information embedded in the structure of the semantic space:

---

**Algorithm 1** - Unsupervised disambiguation algorithm for term $t$

---
1 : Build the cohort of terms with similarity to $t$ above $\sigma_{top}$. There are $N$ such terms.
2 : Form $N$ clusters by assigning its own cluster to each of the $N$ terms and setting cluster vectors identical to term vectors
3 : Compute initial intercluster similarity matrix of size $N \times N$
4 : $K \leftarrow N$
5 : **repeat**
6 :    Merge the two closest clusters by assigning the weighted sum of their centroid vectors to one cluster and discarding the other
7 :    $K \leftarrow K - 1$
8 :    Compute intercluster similarity matrix of size $K \times K$
9 :    Valid disambiguations are groups of clusters where the largest element in the matrix is under threshold $\sigma_{threshold}$
10 : **until** $K = 1$

---

A disambiguation is a set of senses and each sense is a vector in the semantic space. This vector will often be used by itself, for example by replacing in the query vector the original vector of the word before disambiguation. It can be visualized by listing a cohort of its neighboring terms, as in the examples below.

A reasonable value for the similarity $\sigma_{top}$ is $0.40$, so that closely related terms are included but the cohort size $N$ is still limited. For $\sigma_{threshold}$, a value of about three standard deviations, i.e. $0.175$, has been empirically chosen, both well below any significant similarity measure and well above the noise discussed above.

In what follows, only groups of four, three and two clusters will be considered.

## 4. Disambiguation examples

Since a cohort above $\sigma_{top}$ is never very large (a cohort is usually of length 10 to 100) the process is very fast. A list of disambiguations for the 21,365 terms which occur between 5 and 5,000 times in the patent database builds in a few hours on a low-performance desktop machine running a single thread. Out of the total, 411 words were found to have four or more senses, 1010 words had three senses and 4130 words had only two senses.

Following the case of the word *mantle* (Example 1), it is seen that when there are four clusters, i.e. when *mantle* has four senses, it resolves into four distinct vectors associated with the top words *stirrer*, *burkitt*, *four-mouth* and *sieve*; these vectors have similarities of $0.639$, $0.493$, $0.472$ and $0.402$ to the parent word *mantle*. At this clusterization level, senses #1 and #3 are both referring to containers; their similarity is $0.391$, well above $\sigma_{threshold}$ and the disambiguation is rejected.

Next, clusters #1 and #3 are merged, resulting in three senses with the top words *stirrer*, *burkitt* and *sieve* and similarities $0.649$, $0.493$ and $0.402$ to the parent. Senses #1 and #3 are still too close (similarity too high at $0.259$) and the disambiguation is rejected.

Finally, in the last step, the two final clusters have a similarity of only $0.096$. The algorithm stops and only a two-sense disambiguation is retained.

> Mantle
> - sense #1 (0.639)  |  stirrer, four-neck, three-neck, four-necked, gas-adapter, equipped, thermometer …
> - sense #2 (0.493)  |  burkitt, b-lymphoblastic, no-hodgkin, lymphoplasmacytic, hodgkin, lymphoma, mediastinal …
> - sense #3 (0.472)  |  four-mouth, one-liter, mantle, gas-adapter, two-liter, inclined-blade, axially-conveying …
> - sense #4 (0.402)  |  sieve, nanoparticle-loaded, silicate-alumina-based, silica-alumina-based, post-calcined …
>
> - sense #1 (0.649)  |  stirrer, four-neck, three-neck, four-necked, gas-adapter, equipped, thermometer …
> - sense #2 (0.493)  |  burkitt, b-lymphoblastic, no-hodgkin, lymphoplasmacytic, hodgkin, lymphoma, mediastinal …
> - sense #3 (0.402)  |  sieve, nanoparticle-loaded, silicate-alumina-based, silica-alumina-based, post-calcined …
>
> - sense #1 (0.655)  |  stirrer, four-neck, three-neck, four-necked, gas-adapter, equipped, thermometer …
> - sense #2 (0.493)  |  burkitt, b-lymphoblastic, no-hodgkin, lymphoplasmacytic, hodgkin, lymphoma, mediastinal …
>
> Example 1 - The only valid disambiguation is the last one, with two vectors having similarities $0.665$ and $0.493$ to the parent term *mantle*. Depending on the context, vector #1 or vector #2 will replace the parent vector in a query. For convenience, a list of the terms closest to each vector is also given. *Mantle* occurs 2119 times in the database, *burkitt* occurs 242 times and has no disambiguation.

The word *pig* is another example where direct examination of the similarity matrix of the word's cohort yields an obvious disambiguation:

> Pig
> - sense #1 (0.726)  |  guinea, rabbit, goat, donkey, sheep, monkey, rodentia …
> - sense #2 (0.541)  |  pipeline, pipeline-based, over-land, performance-to-power, selected-stage, remaining-stage …
>
> Example 2 - Pig refers of course to an animal (sense #1) but also to a *pipeline inspection gauge* in the oil industry (sense #2). There are other technical senses for *pig* but they are not detected in the present setup, despite the fact that *pig*, which occurs 3079 times, is indeed associated with *picture-in-guide* which occurs 17 times in the database.

It would be far more difficult to unravel by simple inspection and without prior knowledge the senses of the word *bather*:

| Bather |
|---|
| • sense #1 (0.348) \| spa, un-configured, partial-region, proration, water-circulation, hot-tub, fc-hinge … <br> • sense #2 (0.303) \| sic-schottky, schottky, second-barrier, source-gated, first-barrier, jbs, recessed-drain … <br> • sense #3 (0.301) \| fixture, forbidden-region, visible-light-communication, strip-style, in-grade, half-thru, grazer … <br> • sense #4 (0.288) \| ion-implantation, chemo-mechanical, implanting, doping, plad, low-doped, ion-implanting … |
| • sense #1 (0.373) \| fixture, forbidden-region, visible-light-communication, strip-style, in-grade, half-thru, grazer … <br> • sense #2 (0.348) \| spa, un-configured, partial-region, proration, water-circulation, hot-tub, fc-hinge … <br> • sense #3 (0.303) \| sic-schottky, schottky, second-barrier, source-gated, first-barrier, jbs, recessed-drain … |
| • sense #1 (0.440) \| spa, un-configured, partial-region, proration, water-circulation, hot-tub, fc-hinge … <br> • sense #2 (0.373) \| fixture, forbidden-region, visible-light-communication, strip-style, in-grade, half-thru, grazer … |
| Example 3 - In this case, diambiguations are significant at levels 4, 3 and 2. At level #4, the most obvious interpretation is the first one: a *bather* in a *spa*; the second interpretation refers to a *Schottky **bather** diode*, occurring in a number of patents; the third is possibly a misspelling (frequent in the patent database); it may also be due to the fact that the word *bather* occurs in several unrelated contexts. Sense #4 refers to inventor Wayne Bather, who is an author of a number of patents relative to *ion implantation*. |

*Rage*, which occurs 432 times, is another interesting example; except to specialists of very specific subjects, the several senses of *rage* are not obvious, but they are quite relevant for query disambiguation.

| Rage |
|---|
| • sense #1 (0.524) \| pity, rage, remorse, pride, outrage, walser, veteran … <br> • sense #2 (0.370) \| esrage, srage, aortopathies, albumin, oncomodulin, anti-cardiolipin, serum … <br> • sense #3 (0.356) \| ramasamy, intelligent, acrylate-alkyl, rage, rosin, disproportionated, no-lighting-system … <br> • sense #4 (0.340) \| inhibits, inhibitor, angiogenesis, nf-kb, inhibitory, antagonizes, competitively … |
| • sense #1 (0.593) \| pity, rage, remorse, pride, outrage, walser, veteran … <br> • sense #2 (0.370) \| esrage, srage, aortopathies, albumin, oncomodulin, anti-cardiolipin, serum … <br> • sense #3 (0.340) \| ramasamy, intelligent, acrylate-alkyl, rage, rosin, disproportionated, no-lighting-system … |
| • sense #1 (0.649) \| pity, rage, remorse, pride, outrage, walser, veteran … <br> • sense #2 (0.340) \| esrage, srage, aortopathies, albumin, oncomodulin, anti-cardiolipin, serum … |
| Example 4 - The first sense is the common understanding of the word. The second sense corresponds to the *receptor for advanced glycation end-products (RAGE)*, related to soluble forms of RAGE (sRAGE), including the splice variant *endogenous secretory RAGE (esRAGE)*, occurring in a number of patents. *Ramasamy* is the name of a widely cited author of articles on RAGE and ischemic injury. Finally, RAGE and angiogenesis occur simultaneously in a number of patents. |

At level 2 (two senses), all three medical senses have been conflated, as expected. The intercluster distances at levels 4, 3 and 2 are respectively 0.151, 0.160 and 0.100, illustrating the fact that the smallest intercluster distance is not necessarily uniformly decreasing in agglomerative hierarchical clustering, depending on the linkage, but inversions are without consequence in this application.

*Elephant* (occurring 465 times) is another example of unexpected disambiguation for a non-specialist:

| Elephant |
|---|
| • sense #1 (0.331) \| forwarding, no-management, no-first-hop, bit-indexed, loop-detect, no-designation … <br> • sense #2 (0.324) \| less-congested, sideways-directed, expellant, cost-cognitive, costs, preconfigures, congested … <br> • sense #3 (0.315) \| feeley, sigops, sigcomm, elephant, dodecanol, review, paral … |
| • sense #1 (0.426) \| less-congested, sideways-directed, expellant, cost-cognitive, costs, preconfigures, congested … <br> • sense #2 (0.331) \| sub-label, forwarding, no-management, no-first-hop, bit-indexed, loop-detect, no-designation … |
| Example 5 - Some disambiguations are unexpected, but quite useful. In computer networking, an *elephant flow* is an extremely large (in total bytes) continuous flow; its opposite is a *mouse flow*. Sense #3 is still a bit ambiguous, as Michael Feeley is a widely cited author on computers and operating systems, and *dodecanol* is a compound related to a sex pheromone of female Asian elephants … |

## 5. Conclusions and future work

In this work, we have presented a simple algorithm for the unsupervised disambiguation of terms in a semantic space without *a priori* training. It is expected that this algorithm should work with any kind of embedding, provided meaningful similarities between words can be defined, i.e. provided the embedding fulfills at least two conditions, (i) words which close meanings should have high similarities and (ii) words with unrelated meanings should have near-zero similarities, so that $\sigma_{threshold}$ can be significantly lower than $\sigma_{top}$. However, the exact nature of the embedding algorithm [13, 20] may have a substantial impact on its usefulness for database querying and disambiguation and we'll compare in a forthcoming article the disambiguation and querying performances of random vectors and of neural networks such as Mikolov's *et al.* [6, 7].

The ability to quickly and automatically distinguish word senses in a semantic space built "on the fly" from constantly evolving data sets such as Wikipedia, from repositories of patent grants and applications [19] or from large sets of documents in Technology Assisted Review and e-discovery [21, 22] opens the way to a number of useful applications.

On the practical side, query disambiguation, i.e. automatic selection of the best sense for each query word, should substantially enhance search precision, resulting in fewer, better focused documents and reducing analysis time and cost.

More fundamentally, while (as indicated in the introduction) disambiguation is understood here to mean sense *induction*, it could conceivably be combined with information from items such as Wikipedia's disambiguation pages or Wiktionary to automatically align with or even create sense *definitions* and, perhaps with some human supervision, generate evolving, continually up-to-date thesauri and dictionaries.

Unsupervised disambiguation would also be a very useful tool for the systematic exploration of semantic spaces, both in diachrony and in synchrony. For example, how do word senses evolve from year to year in a newspaper? in Wikipedia? or, how do word senses compare in a given year between a newspaper? a tabloid? Wikipedia? TV and radio news?

We intend to investigate some of these questions in our future research, possibly including non-English data sets.

## Bibliography

[1] Yarowsky, D., Unsupervised word sense disambiguation rivaling supervised methods, *Proc. of the 33rd Annual Meeting of the Association for Computational Linguistics*, 1995.

[2] See for example the page Word-sense disambiguation in Wikipedia.

[3] Iacobacci, I., Pilehvar, M. T. and Navigli, R., Embeddings for Word Sense Disambiguation: An Evaluation Study, *Proceedings of the 54th Annual Meeting of the Association for Computational Linguistics, pp. 897-907*, 2016.

[4] Butnaru A.M., Ionescu R.T. and Hristea, F., ShotgunWSD: An unsupervised algorithm for global word sense disambiguation inspired by DNA sequencing, *arXiv:1707.08084v1 [cs.CL] 25 July 2017* .

[5] Fellbaum, C., editor, WordNet: An Electronic Lexical Database, *MIT Press*, 1998.

[6] Mikolov, T., Chen, K., Corrado, G. and Dean, J., Efficient Estimation of Word Representations in Vector Space, *arXiv:1301.3781v3*, 2013.

[7] Mikolov, T., Sutskever, I., Chen, K., Corrado, G. and Dean, J., Distributed Representations of Words and Phrases and their Compositionality, *arXiv:1310.4546v1*, 2013.

[8] Yates, A. and Etzioni, O., Unsupervised Methods for Determining Object and Relation Synonyms on the Web, *Journal of Artificial Intelligence Research 34 (2009) pp. 255-296*, 2009.

[9] Firth, J. R., A Synopsis of Linguistic Theory, 1930-1955, *In Studies in Linguistic Analysis, Special volume of the Philological Society, Oxford, UK*.

[10] Salton, G., editor, The SMART retrieval system: Experiments in automatic document processing, *Prentice-Hall, Inc., 1971*.

[11] Salton, G., Automatic Text Processing, *Addison-Wesley Publishing Company*, ISBN 0-201-12227-8, 1989.

[12] Deerwester, S. *et al.*, Improving Information Retrieval with Latent Semantic Indexing, *Proceedings of the 51st Annual Meeting of the American Society for Information Science (25)*, 1988.

[13] Sahlgren, M., The Word-Space Model: Using Distributional Analysis to Represent Syntagmatic and Paradigmatic Relations between Words in High-dimensional Vector Spaces, *PhD Dissertation*, Stockholm University, Sweden, 2006.

[14] Widdows, D. and Cohen, T., The Semantic Vectors Package: New Algorithms and Public Tools for

Distributional Semantics , *Fourth IEEE International Conference on Semantic Computing (IEEE ICSC2010)*, 2010.

[15] Levy, O. and Goldberg, Y., Dependency-Based Word Embeddings, Short paper in ACL 2014.

[16] QasemiZadeh, B. and Handschuh, S., Random Indexing Explained with High Probability, *Proceedings of the 18th International Conference on Text, Speech and Dialog*, Springer International Publishing, 2015.

[17] Dasgupta S., Technical Perspective: Strange Effects in High Dimension, *Communications of the ACM, Vol. 53* (2010) No. 2, Page 96.

[18] Delpech, J.-F and Ploux, S., Random vector generation of a semantic space, *arXiv:1703.02031*, 2017.

[19] Delpech, J.-F., Semantic Technology-Assisted Review (STAR): Document analysis and monitoring using random vectors, *arXiv:1711.10307*, 2017.

[20] Melamud, O., McClosky, D., Patwardhan, S. and Bansal, M., The Role of Context Types and Dimensionality in Learning Word Embeddings, *arXiv:1601.00893v2*, 2017.

[21] Grossman, M. R. and Cormack, G. V., Technology-Assisted Review in Electronic Discovery , *Chapter to be published in Ed Walters (ed.), Data Analysis in Law (Taylor & Francis Group, forthcoming 2018)* .

[22] Cormack, G. V. and Grossman, M. R. , Evaluation of Machine-Learning Protocols for Technology-Assisted Review in Electronic Discovery , SIGIR'14, July 6-11, 2014, ACM 978-1-4503-2257-7/14/07.